\documentclass{article}
\usepackage{amsmath}
\usepackage{amsfonts}
\usepackage{graphicx}
\usepackage{tikz}
\usepackage[english]{babel}
\usepackage{amsthm}
\usepackage{amsmath, amsfonts, amssymb}
\theoremstyle{definition}
\newtheorem{definition}{Definition}[section]
\usepackage{epstopdf} 
\usepackage{algorithm}
\usepackage{algpseudocode}
\usepackage{url}
\usepackage{subcaption}  
\usepackage{booktabs} 
\usepackage{multirow} 
\usepackage{blindtext}
\usepackage[a4paper, total={6in, 8in}]{geometry}
\usepackage{epstopdf}

\usepackage{authblk}

\title{Spatial-Temporal Bearing Fault Detection Using Graph Attention Networks and LSTM}

 \author[1]{Moirangthem Tiken Singh}
 \author[2]{Rabinder Kumar Prasad}
 \author[3]{Gurumayum Robert Michael}
 \author[4]{N. Hemarjit Singh}
 \author[5]{N. K. Kaphungkui}

 \affil[1,2]{Department of Computer Science and Engineering, DUIET, Dibrugarh University}
 \affil[3,4,5]{Department of Electronics and Communication Engineering, DUIET, Dibrugarh University}

 \affil[1]{Email: tiken.m@dibru.ac.in}
 \affil[2] {Email: rkp@dibru.ac.in}
 \affil[3]{Email: robertmichael@dibru.ac.in}
 \affil[4]{Email: nhsingh@dibru.ac.in}
 \affil[5]{Email: pipizs.kaps@gmail.com}

\date{}

\begin{document}

\maketitle
\begin{abstract}

\noindent \textbf{Purpose:} This paper aims to enhance bearing fault diagnosis in industrial machinery by introducing a novel method that combines Graph Attention Network (GAT) and Long Short-Term Memory (LSTM) networks. This approach captures both spatial and temporal dependencies within sensor data, improving the accuracy of bearing fault detection under various conditions.

\noindent \textbf{Methodology:} The proposed method converts time series sensor data into graph representations. GAT captures spatial relationships between components, while LSTM models temporal patterns. The model is validated using the Case Western Reserve University (CWRU) Bearing Dataset, which includes data under different horsepower levels and both normal and faulty conditions. Its performance is compared with methods such as K-Nearest Neighbors (KNN), Local Outlier Factor (LOF), Isolation Forest (IForest) and GNN-based method for bearing fault detection (GNNBFD).

\noindent  \textbf{Findings:} The model achieved outstanding results, with precision, recall, and F1-scores reaching 100\% across various testing conditions. It not only identifies faults accurately but also generalizes effectively across different operational scenarios, outperforming traditional methods.

\noindent  \textbf{Originality:} This research presents a unique combination of GAT and LSTM for fault detection, overcoming the limitations of traditional time series methods by capturing complex spatial-temporal dependencies. Its superior performance demonstrates significant potential for predictive maintenance in industrial applications.
\end{abstract}


\section{Introduction}

Mechanical equipment is an essential part of modern industries because of advancements in science and technology, making health monitoring technologies critical for ensuring system integrity. The reliability and safety of industrial machinery depend heavily on rotating components, making condition monitoring and fault diagnosis necessary for maintaining operational efficiency and preventing failures \cite{randall2021vibration}. Rolling bearings are critical components in rotating machinery and are prone to failure because of high-speed motion, heavy loads, and exposure to high temperatures. Such failures can directly affect machine performance, leading to safety risks and costly maintenance. Industry data shows that defects in rolling bearings account for 30-40\% of failures in rotating machinery \cite{tandon1999review}.

\noindent Bearing fault diagnosis has progressed through three stages: manual experience-based approaches, signal processing-based methods, and AI-driven techniques. Traditional methods work well for diagnosing simple systems with singular faults but struggle with complex systems involving multiple faults or unpredictable failures \cite{liu2018artificial}. As machinery grows more complex, traditional approaches are becoming inadequate \cite{abid2021review}. The availability of large amounts of operational data from sensors has paved the way for AI techniques like deep learning, which have shown promise in extracting fault patterns, improving diagnostic accuracy, and reducing costs \cite{liu2016rolling}. However, deep learning models often assume data independence, limiting their ability to capture interdependencies crucial for early fault detection \cite{zhang2020deep}.

\noindent Traditional time series analysis methods cannot capture both local and global dependencies, particularly in non-linear, irregular, and multi-scale temporal data. They also overlook the spatial structure present in different segments. This paper introduces a cutting-edge fault diagnosis approach combining Graph Attention Networks (GAT) \cite{velivckovic2017graph} and Long Short-Term Memory (LSTM) networks \cite{hochreiter1997long} to address these challenges.

\noindent The approach enhances fault detection by capturing both spatial relationships between components and temporal dynamics in sensor data. A graph-based representation of time series data is constructed using entropy-based segmentation and Dynamic Time Warping (DTW) \cite{liu2024novel} to compute segment similarities. The GAT captures spatial dependencies, while LSTM models temporal patterns. The attention-based neural network model improves the accuracy, reliability, and interpretability of fault detection, focusing on detecting complex systems like bearing faults.

The objectives are:

\begin{itemize}

    \item Develop a graph-based representation of time series data by segmenting with entropy and computing segment similarities using DTW.

    \item Design an attention-based neural network model using GAT for spatial dependencies and LSTM for temporal dynamics.

    \item Optimize segmentation and graph construction by determining the ideal window size for entropy-based segmentation to minimize noise and computational complexity.

\end{itemize}

\section{Literature Review}

Extensive research has been conducted on intelligent bearing fault diagnosis, a key aspect of predictive maintenance and condition monitoring. Early machine learning methods, such as Principal Component Analysis (PCA) \cite{gu2018fault}, Support Vector Machines (SVM) \cite{hwang2015support}, and K-Nearest Neighbor (KNN) \cite{kumar2023fault}, achieved notable success in fault classification. These models significantly improved classification accuracy over traditional signal processing techniques by utilizing statistical and geometric approaches \cite{zhang2020deep}. However, classical algorithms struggle with high-dimensional data and often fail to capture the underlying nonlinearities in noisy, complex environments, limiting their effectiveness in varying operational conditions \cite{lee2018machine}.

\noindent A significant limitation of traditional models lies in their difficulty in modeling intricate nonlinear interactions between input features, such as time-domain and frequency-domain signals, and output labels (fault categories) \cite{gholaminejad2023comparative}. Despite their initial success, shallow learning models frequently underperform in environments with fluctuating operating settings or significant noise \cite{ghorvei2021unsupervised}.

\noindent Deep learning, on the other hand, offers a breakthrough solution. As an advanced subset of machine learning, deep learning can model complex nonlinearities through multiple layers of abstraction, enabling it to handle high-dimensional data and uncover subtle patterns that traditional models often overlook \cite{kim2022failure}. Techniques such as Convolutional Neural Networks (CNNs) \cite{freire2024fault} and Recurrent Neural Networks (RNNs) \cite{liao2024multi} have demonstrated great efficacy in extracting features from raw vibration data and capturing temporal dependencies, making them particularly well-suited for fault diagnosis \cite{mahesh2024data}. However, CNNs often require large labeled datasets for training, which can be a significant limitation when data is scarce or expensive to acquire. Janssens et al. \cite{janssens2016convolutional} were pioneers in applying CNNs to bearing fault diagnosis by leveraging the spatial structure of data to capture covariance in frequency decomposition from accelerometer signals. While this approach was innovative, it is computationally expensive and prone to overfitting with small datasets. Guo et al. \cite{guo2016hierarchical} refined this by incorporating adaptive learning rates and momentum components to balance training speed and accuracy, though even this enhancement struggles with noisy and imbalanced datasets, affecting robustness. Xia et al. \cite{xia2017fault} further improved CNN training by integrating temporal and spatial information from multi-sensor data, but their method still requires extensive pre-processing and domain expertise to achieve high accuracy.

\noindent Zhang et al. \cite{zhang2019deep} developed a method to process vibration signals of varying sequence lengths using residual learning and 1D convolutional layers for precise feature extraction. While this approach improved accuracy, the model’s complexity increased training time and computational costs, limiting its practicality for real-time applications. Meng et al. \cite{meng2019data} advanced this by combining deep convolutional networks with residual learning, which enhanced diagnostic accuracy even with limited training data, though the model remains vulnerable to overfitting in imbalanced datasets. Zhang et al. \cite{zhang2018fault} employed a deep fully convolutional neural network (DFCNN) to transform vibration signals into images for improved input handling. However, converting time-series data into images can result in information loss and increased model complexity. Choudhary et al. \cite{choudhary2021convolutional} explored thermal imaging alongside CNNs for diagnosing fault conditions in rotating machinery, but incorporating thermal data adds complexity, and accurate thermal imaging is expensive and sensitive to environmental conditions. Xu et al. \cite{xu2022fault} proposed the Online Transfer Convolutional Neural Network (OTCNN), leveraging pre-trained CNNs and source domain features to adapt to real-time data, enhancing the accuracy of rolling bearing fault diagnoses. Although effective, OTCNN is susceptible to domain shift when the source and target domains differ significantly.

\noindent Shao et al. \cite{shao2017novel} employed maximum correlation entropy as a loss function in a deep autoencoder (DAE), optimizing key parameters using the artificial fish swarm algorithm to better match signal characteristics. Despite these improvements, this model remains sensitive to hyperparameter selection and may not generalize well across different fault types. Mao et al. \cite{mao2021new} introduced a novel loss function incorporating a discriminant regularizer and symmetric relation matrix to capture structural discriminant information for fault types. However, the increased complexity of the model lengthens training time, and tuning the regularizer remains challenging. Hao et al. \cite{hao2024novel} employed an ensemble deep autoencoder (EDAE) with Fourier transforms as input for fault diagnosis. While Fourier transforms help isolate frequency-domain features, they may overlook time-domain information, leading to incomplete representations of fault characteristics. Similarly, Liu et al. \cite{liu2024intelligent} transformed sensor data into time-frequency representations (TFRs) using continuous wavelet transforms and fed them into an autoencoder enhanced by a Wasserstein generative adversarial network. This approach effectively handles nonstationary signals but is computationally expensive and sensitive to the choice of wavelet functions. Li et al. \cite{li2024intelligent} utilized a deep convolutional autoencoder (DAE) with a modified loss function to analyze wavelet transmissibility data, but wavelet-based methods are computationally intensive and may struggle to capture complex temporal dependencies. Qu et al. \cite{qu2024mechanical} proposed a semi-supervised learning method based on a variational autoencoder (VAE), which improved classification performance with limited labeled data by leveraging the VAE’s generative capabilities. However, VAEs can experience mode collapse, where the model fails to generate diverse outputs, limiting generalization. Finally, Chang et al. \cite{chang2024rolling} optimized the integration of variational mode decomposition (VMD) and a stacked sparse autoencoder (SSAE) using the Dung Beetle Optimization (DBO) algorithm. While this approach shows promise, the DBO algorithm is computationally expensive and may converge slowly in high-dimensional spaces.

\noindent Chen and Li \cite{chen2017multisensor} were pioneers in utilizing deep belief networks (DBNs) for bearing fault diagnosis by combining them with stacked autoencoders (AE) through a multi-sensor feature fusion approach. Although their method improved fault detection accuracy, DBNs generally suffer from slow training times and sensitivity to the initialization of network weights. Jin et al. \cite{jin2023bearing} proposed a strategy that integrates VMD, feature extraction (FE), and an enhanced DBN. Their approach processes noisy vibration signals with VMD, extracts key features, and employs an improved butterfly optimization algorithm to fine-tune DBN hyperparameters for precise fault diagnosis. However, while the butterfly optimization algorithm improves accuracy, it remains computationally expensive and may not scale well with large datasets. Pan et al. \cite{pan2023fault} enhanced DBN performance using a free energy sampling method in persistent contrastive divergence (FEPCD), focusing on multi-domain feature extraction to identify fault characteristics. Despite its efficacy, this method increases computational overhead and requires significant domain expertise for feature engineering. Elsamanty et al. \cite{elsamanty2023principal} applied PCA to reduce dimensionality and generate uncorrelated principal components (PCs), preserving most of the data’s variability. These PCs were then used as inputs to a backpropagation neural network (BPNN) for robust fault diagnosis. While PCA helps reduce data complexity, it may discard critical information, potentially limiting diagnostic accuracy in complex cases.

\noindent In addition to CNNs, AEs, and DBNs, several other deep learning methods have been applied to bearing fault diagnosis. Generative adversarial networks (GANs) and their variants have shown promise in tackling bearing fault detection challenges \cite{wang2023adaptive, luo2023intelligent}. Moreover, incorporating Long Short-Term Memory (LSTM) networks has improved RNN performance, yielding successful outcomes in bearing fault diagnosis \cite{zhang2023deep, alkhanafseh2023advanced}. Reinforcement learning has also been applied to boost diagnostic accuracy \cite{wang2023match, kang2023dual}.

\noindent Graph Neural Networks (GNNs) have notably advanced the processing of graph-structured data by integrating convolutional networks, recurrent networks, and deep autoencoders. Xiao et al. \cite{xiao2023graph} introduced a GNN-based method for bearing fault detection (GNNBFD), constructing a graph based on sample similarity. This graph is processed by a GNN, which performs feature mapping, allowing each sample to incorporate information from neighboring nodes, thus enriching its representation. Likewise, the Granger Causality Test-based Graph Neural Network (GCT-GNN) \cite{zhang2024graph} enhances fault detection by organizing time-domain and frequency-domain features into a feature matrix for causal analysis using GNNs.

\noindent In contrast to these methods, the proposed model combines a Graph Transformer Network (GTN) with a Long Short-Term Memory (LSTM) network to capture both temporal and spatial relationships in bearing sensor data. The Graph Transformer efficiently handles spatial dependencies between sensor nodes, while the LSTM captures temporal dynamics, offering a more comprehensive understanding of the data that surpasses existing GNN-based approaches. This combination enables superior fault detection by considering both spatial correlations and temporal patterns in the sensor readings. In the next section, the proposed methodology is illustrated.

\section{Methodology}

\subsection{Data Processing}

\begin{figure}[h!]

    \centering

    \includegraphics[width=0.5\linewidth]{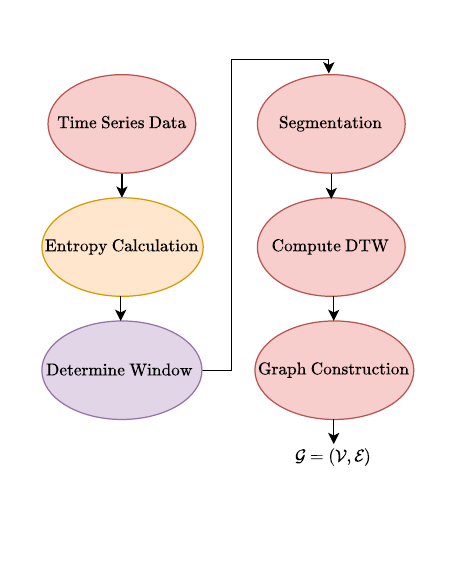}

    \caption{Graph Construction from Time Series Data. Source: Author's own.}

    \label{fig:enter-label}

\end{figure}

This section presents a method that integrates concepts from information theory, time series analysis, and graph theory to construct a graph \(\mathcal{G}\) from time series data. The process involves several key steps: calculating entropy, segmenting the time series, determining the optimal window size, computing Dynamic Time Warping (DTW) distances \cite{liu2024novel}, and constructing a graph based on segment similarities. The entire workflow for generating the graph from time series data is illustrated in Figure \ref{fig:enter-label}.

\noindent To effectively segment the time series data for graph construction, the method uses entropy as a guiding metric. The principle behind this approach is that regions with high entropy naturally serve as boundaries where the time series should be divided into segments. Entropy is computed using Shannon entropy \cite{Fang2009}, defined as follows:

\begin{definition}

    For a discrete random variable \( Z \) with a probability distribution \( P(Z) = \{p_z^1, p_z^2, \dots, p_z^n\} \), the Shannon entropy \( H(Z) \) is given by:

\[ H(Z) = -\sum_{j=1}^{n} p_z^j \log p_z^j \]

where \( p_z^j \) represents the probability of the \( j \)-th outcome.

\end{definition}

\noindent A crucial aspect of the segmentation process is determining the segment length, commonly known as the window size. The chosen window size should capture significant patterns while minimizing the influence of noise. To achieve this, the method aims to find an optimal window size based on entropy information. The entropy \( H(U) \) of a given time series segment \( U \) ensures that the segment is informative and accurately reflects the underlying dynamics of the time series.

\noindent To segment the time series \( T \), it is divided into windows of size \( w \) with a step size \( s \). The entropy for each segment is then calculated, and the average entropy \( \bar{H}(w) \) for each window size is computed as:

\[ \bar{H}(w) = \frac{1}{n(w)} \sum_{i=1}^{n(w)} H(U_i)\]

where \( n(w) \) is the number of segments for window size \( w \), and \( U_i \) is the \( i \)-th segment.

\begin{figure}[h!]

    \centering

    \includegraphics[width=0.5\linewidth]{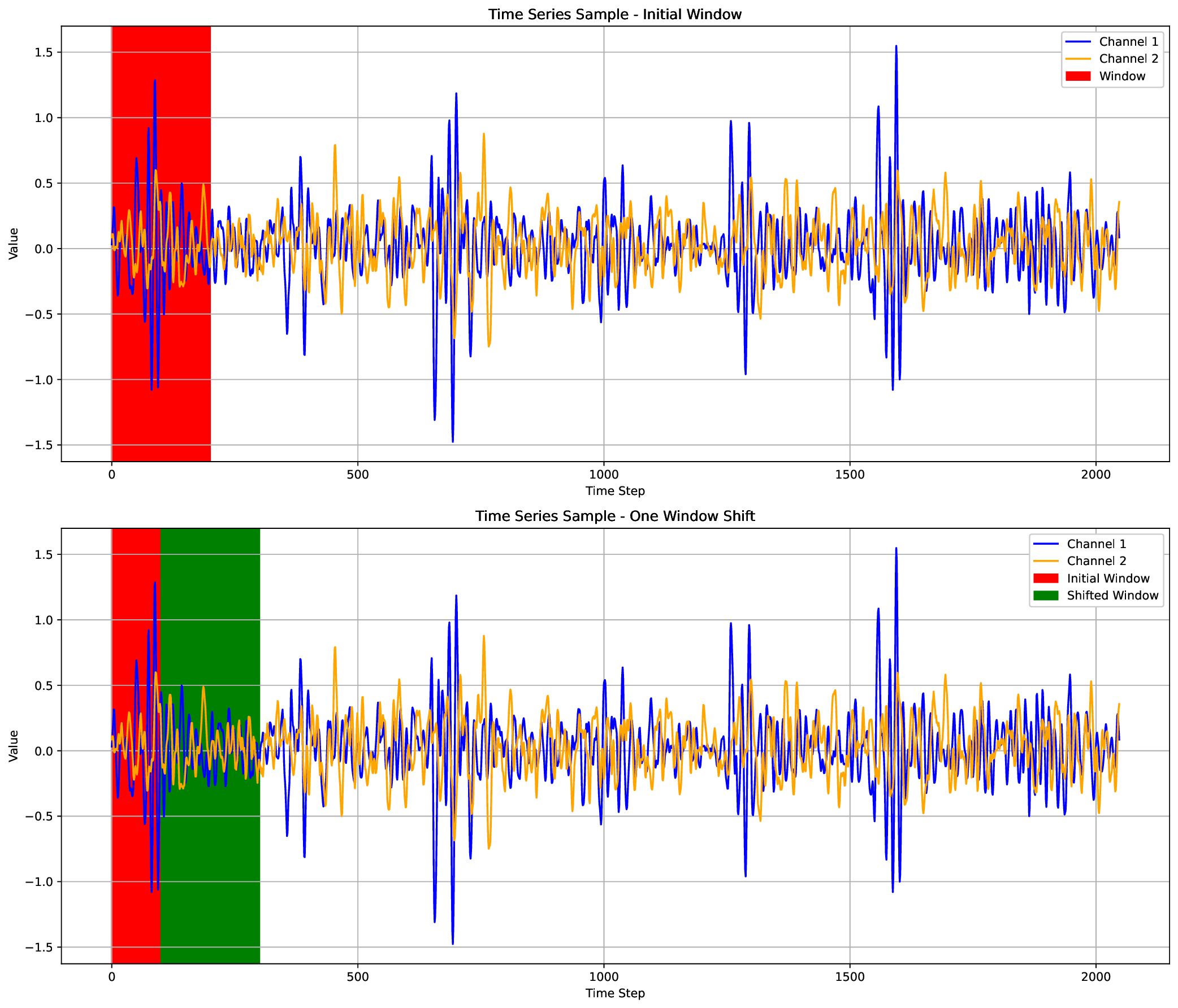}

    \caption{Visualization of Time Series Data with Windowing: The top graph displays two channels of time series data, with the initial time window (highlighted in red) applied. The lower graph illustrates how shifting the window by one step captures and converts the time series data into a graph. Source: Author's own.}

    \label{fig:windows}

\end{figure}

\noindent Although dividing \( T \) into segments of size \( w \) is essential, the challenge lies in determining the optimal value of \( w \). To address this, a notable relationship between entropy and window size has been observed. As Pincus \cite{pincus1991approximate} points out, as \( w \) increases, the number of possible configurations within that window also increases. This leads to a rise in entropy, which may reflect the increased window size rather than any genuine change in the underlying structure of the time series. To make entropy values comparable across different window sizes, the average entropy is normalized by the logarithm of the window size, as defined below:

\[ H_{\text{norm}}(w) = \frac{\bar{H}(w)}{\log w} \]

\noindent This normalization corrects the artificial inflation of entropy associated with larger windows, allowing for a more accurate and fair comparison of the entropy values. This approach better captures the true complexity and dynamics of the time series across various window sizes.

\noindent Using the normalized entropy for each segment and every window size \( w \in \mathcal{W} = \{w_1, w_2, \dots, w_n\} \), our objective is to determine the optimal window size \( w^* \) that maximizes the normalized entropy:

\[ w^* = \arg\max_{w \in \mathcal{W}} H_{\text{norm}}(w) \]

\noindent After determining the optimal window size, the time series is divided into overlapping windows of this size, called segments, as shown in Figure \ref{fig:windows}. This allows for localized analysis of the time series, capturing variations and patterns that might not be apparent in the entire series.

\noindent For a time series \( T \) of length \( N \), with a window size \( w \) and step size \( s \), the segments \( U_i \) are defined as:

\[ U_i = T[i:i+w] \]

for \( i = 0, s, 2s, \dots, N-w \).

\noindent The similarity between segments is then measured using Dynamic Time Warping (DTW), which calculates the minimal distance by warping the time axis to optimally align the series.

\noindent Given two time series \( X = \{x_1, x_2, \dots, x_p\} \) and \( Y = \{y_1, y_2, \dots, y_q\} \), the DTW distance \( D(X, Y) \) is defined as:

\[ D(X, Y) = \min \left(\sum_{k=1}^{K} d(x_{u_k}, y_{v_k}) \right) \]

\noindent where \( d(x_u, y_v) \) is the Euclidean distance between the points \( x_u \) and \( y_v \), and \( (u_k, v_k) \) represents a warping path that aligns the sequences.

\noindent The similarity between two segments \( U_i \) and \( U_j \) is then defined as the inverse of the DTW distance:

\[ \text{Sim}(U_i, U_j) = \frac{1}{1 + D(U_i, U_j)} \]

\noindent The time series segments are represented as nodes in a graph, with edges indicating the similarity between these segments, as measured by DTW. Specifically, a graph \( \mathcal{G} = (\mathcal{V}, \mathcal{E}) \) is constructed, where each node \( v_i \in \mathcal{V} \) corresponds to a segment \( U_i \). An edge \( e_{ij} \in \mathcal{E} \) is added between the nodes \( v_i \) and \( v_j \) if the similarity between the segments \( U_i \) and \( U_j \) exceeds a threshold \( \tau \):

\[ \mathcal{E} = \{(v_i, v_j) \mid \text{Sim}(U_i, U_j) > \tau\} \]

\noindent The weight of each edge represents the degree of similarity between the connected segments.

\subsection{Architecture}
\label{model}

\begin{figure}[h!]

    \centering

    \includegraphics[width=\linewidth]{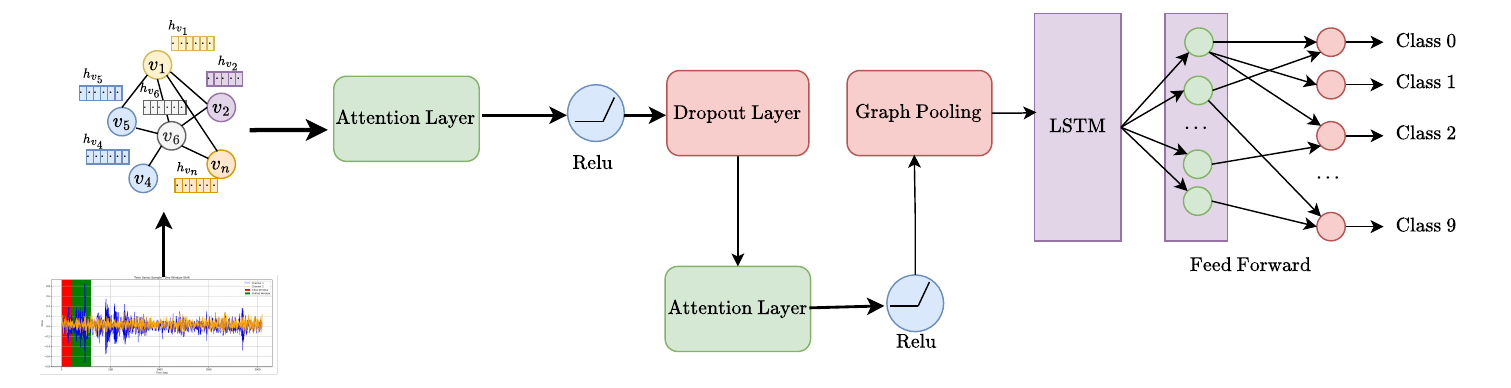}

    \caption{Attention-Based Time Series Analysis Graph Model (ATBTSGM). Source: Author's own.}

    \label{fig:ATBTSGM}

\end{figure}

The proposed method, Attention-Based Time Series Analysis Graph Model (ATBTSGM), is illustrated in Figure \ref{fig:ATBTSGM}. It involves feeding a graph \( \mathcal{G} = (\mathcal{V}, \mathcal{E}) \) into a neural network architecture that leverages an attention mechanism to capture the importance of each node in relation to its neighbors. Each node \( v_i \in \mathcal{V} \) is associated with a feature vector \( \mathbf{h}_i \in \mathcal{R}^{n \times 1} \), representing the node’s relevant attributes. The model employs the attention mechanism from \cite{veličković2018graphattentionnetworks} to compute attention scores between pairs of nodes \( v_i \) and \( v_j \) within a neighborhood.

\begin{figure}[h!]

    \centering

    \includegraphics[width=\linewidth]{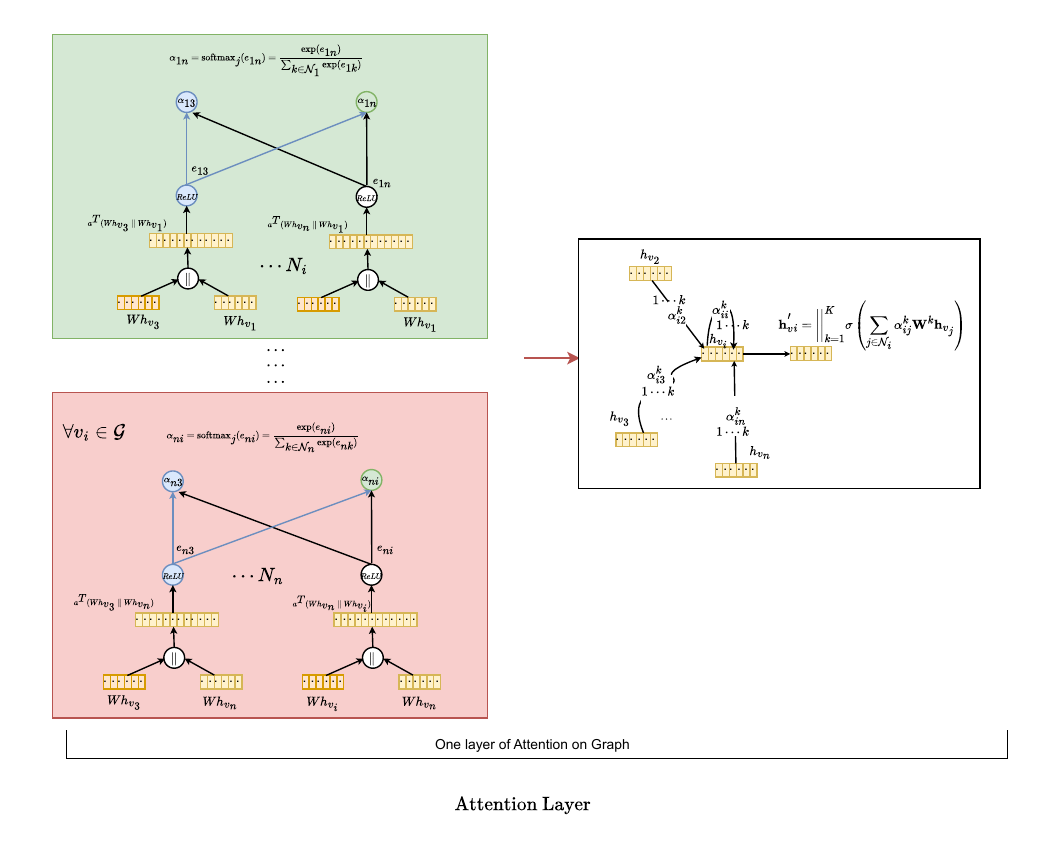}

    \caption{Attention layer in Attention Based Time Series Analysis Graph Model (ATBTSGM). Source: Author's own.}

    \label{fig:attention}

\end{figure}

\noindent The attention mechanism, as illustrated in Figure \ref{fig:attention}, begins by calculating an attention score \( e_{ij} \) between two neighboring nodes based on their feature vectors and learned parameters. The attention score is computed as:

\[e_{ij} = \text{LeakyReLU}\left(\mathbf{a}^T \left[\mathbf{W} \mathbf{h}_i \parallel \mathbf{W} \mathbf{h}_j\right]\right)\]

\noindent where \( \mathbf{a} \) is a learnable weight vector, \( \mathbf{W} \) is a weight matrix applied to the feature vectors, \( \parallel \) denotes the concatenation operation, and LeakyReLU introduces sparsity in the attention scores.

\noindent These raw attention scores \( e_{ij} \) are normalized across all neighbors of a node using the softmax function, yielding the attention coefficients \( \alpha_{ij} \):

\[\alpha_{ij} = \frac{\exp(e_{ij})}{\sum_{k \in \mathcal{N}_i} \exp(e_{ik})}\]

\noindent Here, \( \mathcal{N}_i \) represents the set of neighbors of node \( v_i \), and the softmax normalization ensures that the attention scores sum to one, making them interpretable as probabilities of importance.

\noindent The final representation of each node is obtained by aggregating the feature vectors of its neighbors, weighted by their corresponding attention coefficients:

\[\mathbf{h}_i’ = \sigma\left(\sum_{j \in \mathcal{N}_i} \alpha_{ij} \mathbf{W} \mathbf{h}_j\right)\]

\noindent where \( \sigma \) is a non-linear activation function, such as Exponential Linear Unit (ELU), that introduces non-linearity into the model’s output.

\noindent To improve the model’s robustness, a multi-head attention mechanism is incorporated, where multiple attention heads independently calculate attention coefficients. The results from these heads are concatenated or averaged to form the final output:

\[\mathbf{h}_i’ = \Bigg|\Bigg|_{k=1}^K \sigma\left(\sum_{j \in \mathcal{N}_i} \alpha_{ij}^k \mathbf{W}^k \mathbf{h}_j\right)\]

\noindent This multi-head attention mechanism enables the model to capture diverse aspects of the graph structure, increasing its effectiveness and robustness.

\noindent Here, \( K \) represents the number of attention heads, and this multi-head approach enhances the model’s ability to capture diverse patterns by learning from multiple perspectives simultaneously.

\noindent Following the attention layers, the node-level features are aggregated across the entire graph using a global mean pooling operation. This operation produces a single feature vector representing the entire graph, which is then reshaped to serve as input to a Long Short-Term Memory (LSTM) network. The LSTM captures temporal dependencies in the sequence of graph features by processing the input through its gates: the forget gate, input gate, cell state update, and output gate.

\noindent The hidden state from the final time step of the LSTM encapsulates the learned temporal information and is passed through a fully connected (feedforward) layer for classification. The output of this layer is processed into log-probabilities for each class, ensuring that the model is well-suited for classification tasks by combining the spatial structure of the graph with the temporal dynamics of the data.

\subsection{Algorithms}
Algorithms \ref{alg:time_series_graph} and \ref{alg:attention_lstm} are responsible for encapsulating the entire process described in Section \ref{model}. The main focus of Algorithm \ref{alg:time_series_graph} is on the data processing aspect, whereas Algorithm \ref{alg:attention_lstm} specifically deals with the classification of sensor data.

\begin{algorithm}[!htbp]
\caption{Graph Construction from Time Series Using Entropy and DTW}
\label{alg:time_series_graph}
\begin{algorithmic}[1]
\Require Time series \( T \) of length \( N \), set of window sizes \( \mathcal{W} \), step size \( s \), similarity threshold \( \tau \).
\Ensure Graph \( \mathcal{G} = (\mathcal{V}, \mathcal{E}) \).
\vspace{0.5em}

\State \textbf{Step 1: Entropy Calculation}
\For{each \( w \in \mathcal{W} \)}
    \State Divide \( T \) into windows of size \( w \).
    \State Calculate the average entropy \( \bar{H}(w) \) and normalized entropy \( H_{\text{norm}}(w) \).
\EndFor
\State Choose the optimal window size \( w^* = \arg\max_{w \in \mathcal{W}} H_{\text{norm}}(w) \).

\vspace{0.5em}
\State \textbf{Step 2: Time Series Segmentation}
\State Segment \( T \) into windows of size \( w^* \).

\vspace{0.5em}
\State \textbf{Step 3: DTW Distance Calculation}
\For{each pair of segments \( U_i \) and \( U_j \)}
    \State Compute the DTW distance \( D(U_i, U_j) \) and the similarity score \( \text{Sim}(U_i, U_j) \).
\EndFor

\vspace{0.5em}
\State \textbf{Step 4: Graph Construction}
\State Create a node \( v_i \) for each segment \( U_i \).
\For{each pair \( v_i \) and \( v_j \) where \( \text{Sim}(U_i, U_j) > \tau \)}
    \State Add an edge \( e_{ij} \) to the graph \( \mathcal{G} \) with weight \( \text{Sim}(U_i, U_j) \).
\EndFor

\vspace{0.5em}
\State \Return Graph \( \mathcal{G} \).
\end{algorithmic}
\end{algorithm}

\begin{algorithm}[!htbp]
\caption{Attention-Based Time Series Analysis Graph Model Algorithm}
\label{alg:attention_lstm}
\begin{algorithmic}[1]
\Require Graph \( \mathcal{G} = (\mathcal{V}, \mathcal{E}) \) with node features \( \mathbf{h}_i \).
\Ensure Log-probabilities for classification.
\vspace{0.5em}

\State \textbf{Step 1: Attention Calculation}
\For{each node \( v_i \) in \( \mathcal{V} \)}
    \For{each neighbor \( v_j \in \mathcal{N}_i \)}
        \State Compute attention score:
        \[
        e_{ij} = \text{LeakyReLU}\left(\mathbf{a}^T \left[\mathbf{W} \mathbf{h}_i \parallel \mathbf{W} \mathbf{h}_j\right]\right)
        \]
    \EndFor
    \State Normalize scores:
    \[
    \alpha_{ij} = \text{softmax}_j(e_{ij})
    \]
\EndFor

\State \textbf{Step 2: Feature Aggregation}
\For{each node \( v_i \) in \( \mathcal{V} \)}
    \State Aggregate features:
    \[
    \mathbf{h}_i' = \sigma\left(\sum_{j \in \mathcal{N}_i} \alpha_{ij} \mathbf{W} \mathbf{h}_j\right)
    \]
\EndFor

\State \textbf{Step 3: Multi-Head Attention and Pooling}
\State Apply \( K \) attention heads and global mean pooling:
\[
\mathbf{h}_i' = \Bigg|\Bigg|_{k=1}^K \sigma\left(\sum_{j \in \mathcal{N}_i} \alpha_{ij}^k \mathbf{W}^k \mathbf{h}_j\right)
\]

\State \textbf{Step 4: LSTM and Classification}
\State Pass pooled features through LSTM and classify:
\State \Return Log-probabilities for each class.
\end{algorithmic}
\end{algorithm}

\noindent Algorithm \ref{alg:time_series_graph} constructs the graph \( \mathcal{G} = (\mathcal{V}, \mathcal{E}) \) from time series \( T \) using entropy and DTW. It initially partitions the time series into windows of varying sizes \( w \), and calculates entropy for each segment to determine the optimal window size \( w^* \). It then re-segments the time series using \( w^* \). Subsequently, it obtains similarity scores by calculating DTW distances between all pairs of segments. It represents segments in the graph as nodes and creates edges between nodes that have a similarity score above a threshold \( \tau \). The overall algorithm complexity is \( O(|\mathcal{W}| \cdot N + k^2 \cdot m^2) \), factoring in the length of the time series, number of segments, and segment length.

\noindent Algorithm \ref{alg:attention_lstm} employs an attention mechanism and an LSTM network to process a graph \( \mathcal{G} = (\mathcal{V}, \mathcal{E}) \) with node features \( \mathbf{h}_i \), and generates log-probabilities for classification. It computes attention by calculating scores \( e_{ij} \) for nodes and their neighbors, normalizes them to obtain coefficients \( \alpha_{ij} \), and uses these coefficients to combine the features of nearby nodes. To ensure robustness, it utilizes multihead attention, pools the resultant features, and inputs them into an LSTM to capture temporal relationships. It performs classification by utilizing the LSTM output.

\noindent The attention calculations play a significant role in determining the complexity, resulting in an overall time complexity of \( O(|\mathcal{V}| \cdot |\mathcal{N}_i| \cdot d^2) \), where \( |\mathcal{V}| \) represents the number of nodes, \( |\mathcal{N}_i| \) represents the average number of neighbors, and \( d \) represents the feature dimension.

 \clearpage
\section{Experimental Results }
\subsection{Data Description and Analysis}

\noindent Table \ref{TableData01} summarizes the Case Western Reserve University (CWRU) \cite{CWRU_bearing_data} bearing fault dataset used in this study. This dataset captures various fault conditions at three key locations—Ball, Inner Race, and Outer Race—along with data for a healthy bearing condition (“None”). For each fault location, three fault diameters (0.007, 0.014, and 0.021 inches) were recorded at a sampling frequency of 12 kHz. The datasets, labeled as A, B, and C, correspond to motor loads of 1 hp, 2 hp, and 3 hp, respectively, as detailed in Table \ref{TableData01}.

\begin{table}[htbp!]

\centering

\caption{Description of rolling element bearing datasets at 48 kHz. Source: Author's own.}

\resizebox{\columnwidth}{!}{%

\begin{tabular}{|c|c|c|c|c|}

\hline

\textbf{Fault Location} & \textbf{None} & \textbf{Ball} & \textbf{Inner Race} & \textbf{Outer Race} \\ \hline

\textbf{Category Labels} & 0 & 1, 2, 3 & 4, 5, 6 & 7, 8, 9 \\ \hline

\textbf{Fault Diameter (inches)} & 0 & 0.007, 0.014, 0.021 & 0.007, 0.014, 0.021 & 0.007, 0.014, 0.021 \\ \hline

\textbf{Dataset A (1 hp)} & \multicolumn{4}{c|}{Total Samples: 120832}  \\ \hline

\textbf{Dataset B (2 hp)} & \multicolumn{4}{c|}{Total Samples:  120832} \\ \hline

\textbf{Dataset C (3 hp)} & \multicolumn{4}{c|}{Total Samples: 120832} \\ \hline

\end{tabular}%

}

\label{TableData01}

\end{table}

\noindent To facilitate effective model training, the continuous data stream is segmented into uniform segments based on a predefined sampling rate. This segmentation ensures that all data samples have consistent length, which is essential for training robust models. The stride-based approach creates overlapping segments, enhancing the diversity and quantity of training data without requiring additional raw data. Each segmented sample is assigned a label corresponding to its category or class, a critical step for supervised learning tasks. This segmentation and labeling process improves the model’s ability to learn and generalize.

\begin{figure}[!htbp]

    \centering

    \includegraphics[width=\linewidth]{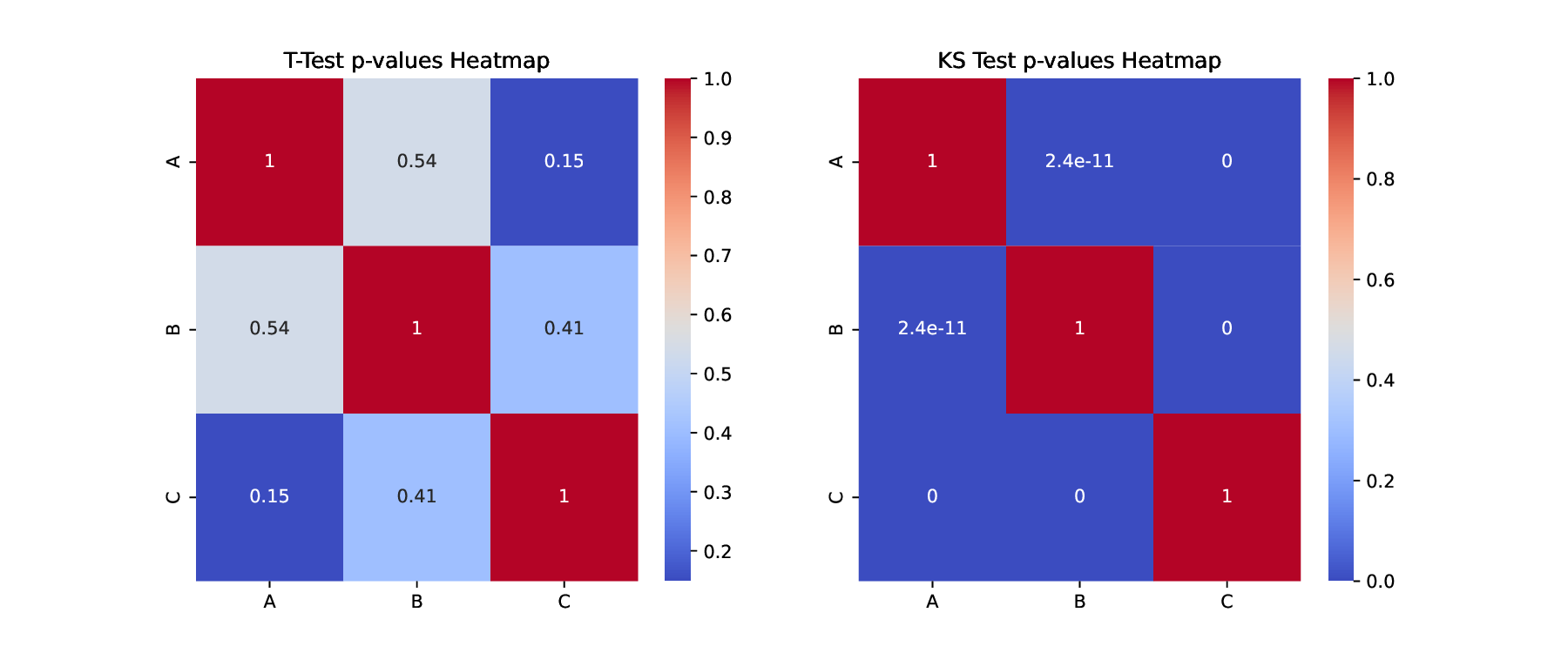}

    \caption{T-Test and Kolmogorov-Smirnov (KS) Test results for Datasets A, B, and C. Source: Author's own.}

    \label{T_test}

\end{figure}

\noindent Each of the three datasets (A, B, and C) contains multidimensional data across 10 distinct classes. To compare these datasets, we conducted statistical tests, including the T-Test and the Kolmogorov-Smirnov (KS) Test, as shown in Figure \ref{T_test}. The T-Test results indicate that there is no statistically significant difference in the means of the datasets. Specifically, the p-values for the comparisons between Dataset A vs B, Dataset A vs C, and Dataset B vs C are 0.5380, 0.1486, and 0.4077, respectively. In contrast, the KS Test reveals significant differences in the distributions of the datasets. For example, the p-value for Dataset A vs B is $2.40 \times 10^{-11}$, while the p-values for Dataset A vs C and Dataset B vs C are both 0.0, indicating substantial divergence between all datasets.

\begin{figure}[h!]

    \centering

    \includegraphics[width=\linewidth]{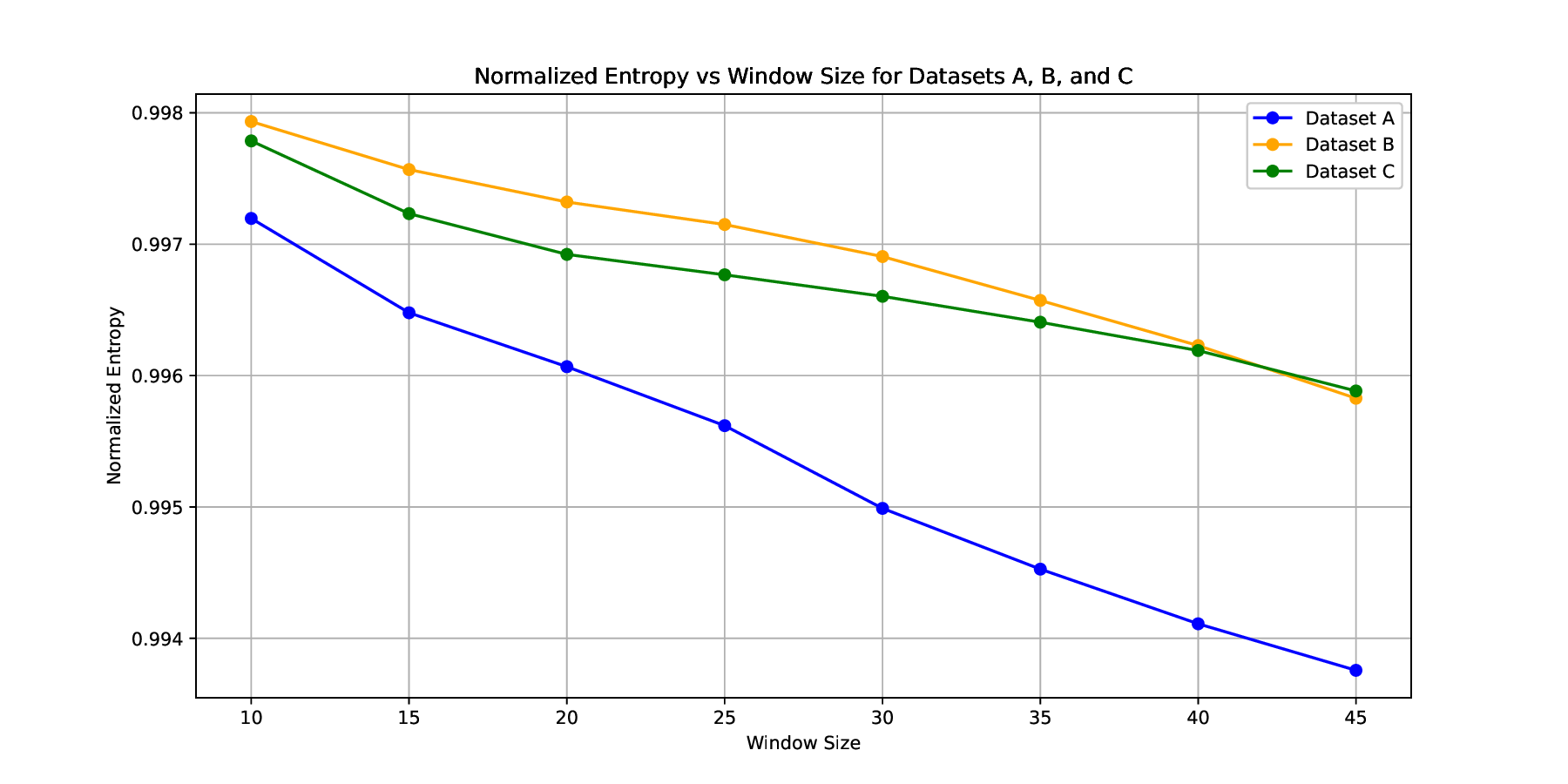}

    \caption{Normalized Entropy versus Window Size for different datasets (A, B, and C). These plots are used to analyze how the entropy of each dataset changes as the window size varies. Source: Author's own.}
    \label{fig:entropy}
\end{figure}

\noindent In this study, an entropy-based method is employed to determine the optimal window size for segmenting the dataset, which is crucial for node creation in graph-based models. Figure \ref{fig:entropy} shows the normalized entropy versus window size for datasets A, B, and C. As the window size increases, entropy declines, signifying reduced randomness or variability in the data, leading to smoother time series representations. By examining these entropy values, an optimal window size range (30 to 40 samples) is identified, balancing noise reduction and data smoothing. This approach ensures the model captures the most relevant features for analysis, such as identifying bearing faults. Thus, entropy serves as a reliable metric for time series segmentation, allowing the model to focus on meaningful patterns in the data while minimizing the effects of noise.

\subsection{Model Evaluation Metrics}

The effectiveness of the fault prediction model in maintaining and diagnosing mechanical systems is evaluated using the following metrics:

\begin{enumerate}

    \item \textbf{Precision} \cite{flach2019performance} measures the proportion of correct positive predictions made by the model, defined as:

    \[     \text{Precision} = \frac{TP}{TP + FP}    \]

    where \( TP \) denotes the number of true positives (correctly predicted faults) and \( FP \) represents false positives (incorrect fault predictions). A high precision score indicates that when the model predicts a fault, it is likely to be accurate, minimizing unnecessary maintenance actions and reducing system downtime and costs.

    \item \textbf{Recall} \cite{flach2019performance} evaluates the model’s ability to correctly identify actual faults, expressed as:

    \[     \text{Recall} = \frac{TP}{TP + FN}    \]

    where \( FN \) denotes false negatives (missed faults). High recall ensures potential issues are detected early, preventing critical system failures and enhancing operational safety.

    \item The \textbf{F1 Score} \cite{flach2019performance} combines Precision and Recall, providing a balanced performance metric, calculated as:

    \[     \text{F1 Score} = 2 \times \frac{\text{Precision} \times \text{Recall}}{\text{Precision} + \text{Recall}}     \]

    This metric is useful when balancing fault detection accuracy and minimizing missed faults is essential.

    \item \textbf{Accuracy (ACC)} \cite{flach2019performance} is the ratio of correctly classified instances to the total number of instances, given by:

    \[    ACC = \frac{TP + TN}{TP + TN + FP + FN}     \]

    where \( TN \) represents true negatives (correctly predicted non-faults).

    \item \textbf{False Alarm Rate (FAR)} \cite{flach2019performance} measures the proportion of false positives out of all negative instances, calculated as:

    \[     FAR = \frac{FP}{FP + TN}     \]

    \item \textbf{Area Under the Curve (AUC)} \cite{flach2019performance} is derived from the Receiver Operating Characteristic (ROC) curve and represents the model’s ability to distinguish between fault and non-fault cases. It is computed as the area under the ROC curve, which plots the trade-off between Recall and FAR.

\end{enumerate}

\noindent These metrics comprehensively assess the model’s performance, including accuracy, fault detection sensitivity, and the ability to minimize false alarms, making them crucial for evaluating the effectiveness of the fault prediction model in mechanical systems.

\subsection{Result Analysis}
For a comprehensive evaluation of the proposed model’s performance, K-Fold cross-validation is employed, a robust technique that ensures generalization and prevents overfitting. The process of K-Fold training involves dividing the dataset into \(K\) subsets or ``folds.'' The model is trained on \(K-1\) folds in iterations, with one fold used for validation each time. This process repeats \(K\) times, with each fold serving as the validation set once. By averaging the final performance across all \(K\) iterations, a more accurate and resilient estimate of the model’s performance is obtained. To accelerate processing speed, the experiments are conducted on a powerful computing system that includes an Intel Xeon processor with 52 cores running at 3.5 GHz, 64 GB of RAM, and a 32 GB graphics card. The model is trained separately on Datasets A, B, and C to examine its generalization ability across different data sources.

\noindent After training, performance is assessed using three key metrics: Precision, Recall, and F1 Score, as summarized in Table \ref{PerformanceTable01}. The model consistently demonstrates strong performance across all datasets, achieving near-perfect results. Specifically, it reaches 99\% Precision, Recall, and F1 Score on both Datasets A and C, showcasing reliable fault detection capabilities. On Dataset B, the model achieves a flawless 100\% across all metrics, indicating robustness and effectiveness in detecting faults under different conditions.

\begin{table}[h!]
\centering
\caption{Performance the Model on Datasets A, B, and C. Source: Author's own.}
\begin{tabular}{|c|c|c|c|}
\hline
\textbf{Dataset}   & \textbf{Precision (\%)} & \textbf{Recall (\%)} & \textbf{F1-Score (\%)} \\ \hline
\textbf{Dataset A} & 99.0                    & 99.0                 & 99.0                   \\ \hline
\textbf{Dataset B} & 100.0                   & 100.0                & 100.0                  \\ \hline
\textbf{Dataset C} & 99.0                    & 99.0                 & 99.0                   \\ \hline
\end{tabular}
\label{PerformanceTable01}
\end{table}

\begin{figure}[h!]
    \centering
    \begin{subfigure}[h!]{0.45\textwidth}
        \centering
        \includegraphics[width=\textwidth]{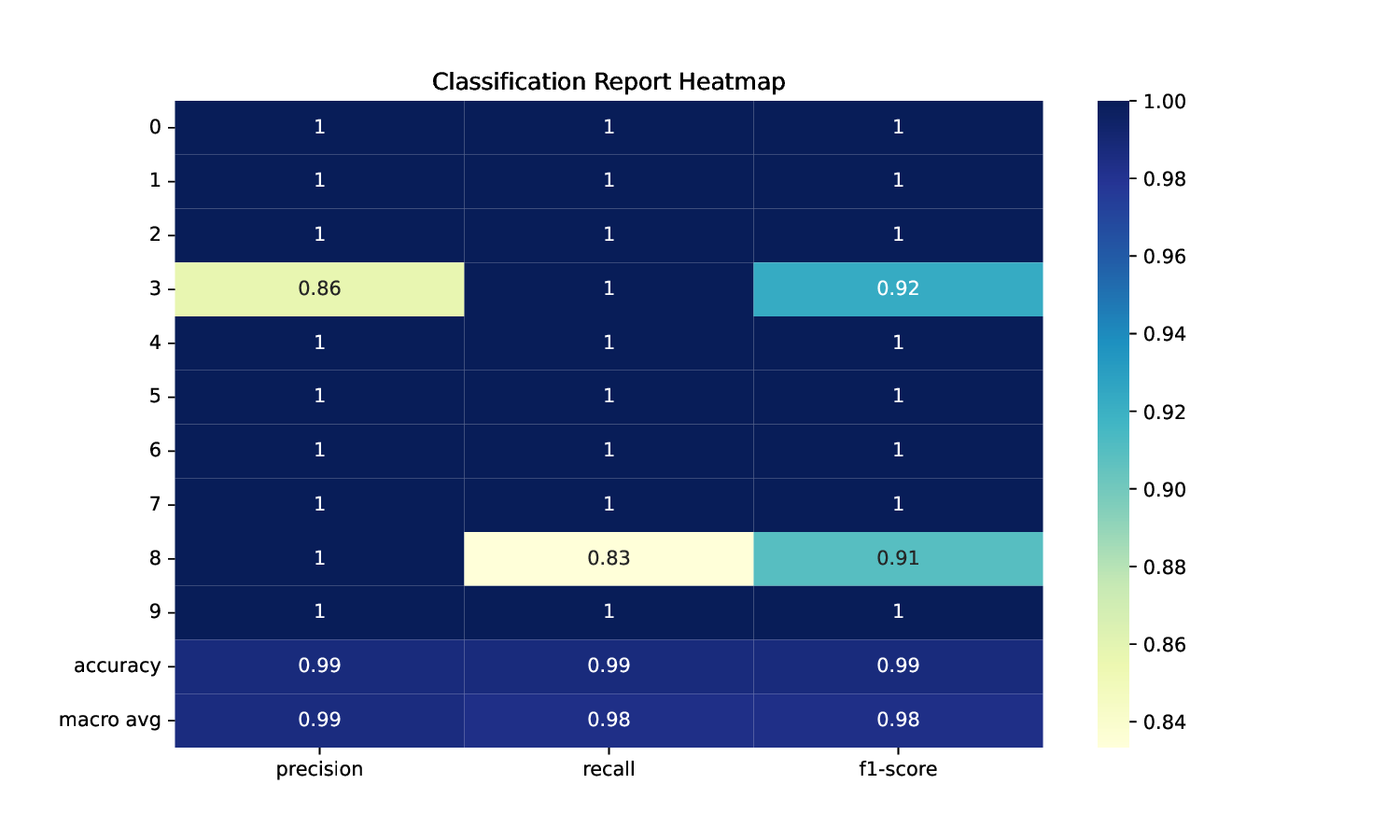}
        \caption{Heatmaps on Dataset A}
        \label{fig:datasetA}
    \end{subfigure}
    \hfill
    \begin{subfigure}[h!]{0.45\textwidth}
        \centering
        \includegraphics[width=\textwidth]{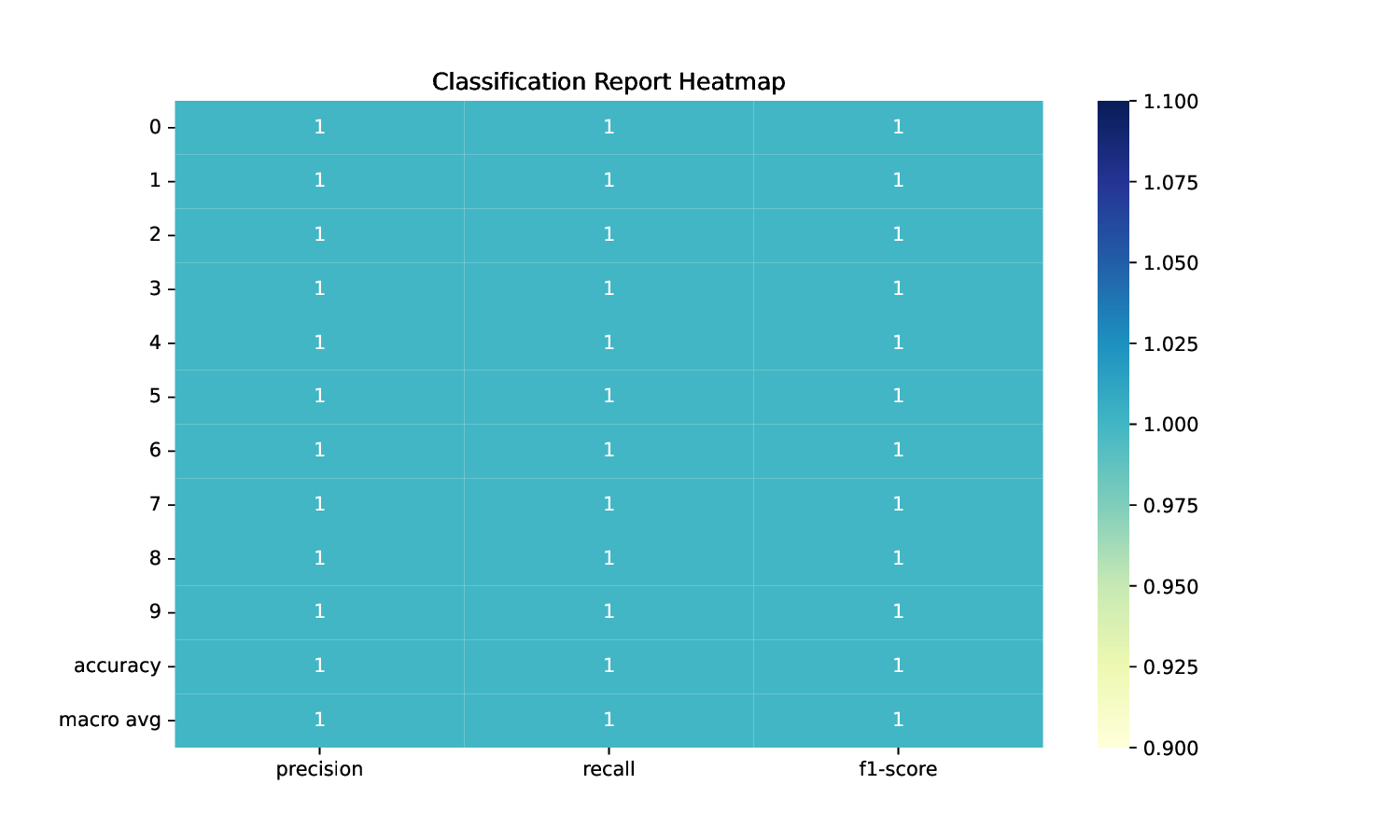}
        \caption{Heatmaps on Dataset B}
        \label{fig:datasetB}
    \end{subfigure}
    \vspace{1em} 
    \begin{subfigure}[h!]{0.5\textwidth}
        \centering
        \includegraphics[width=\textwidth]{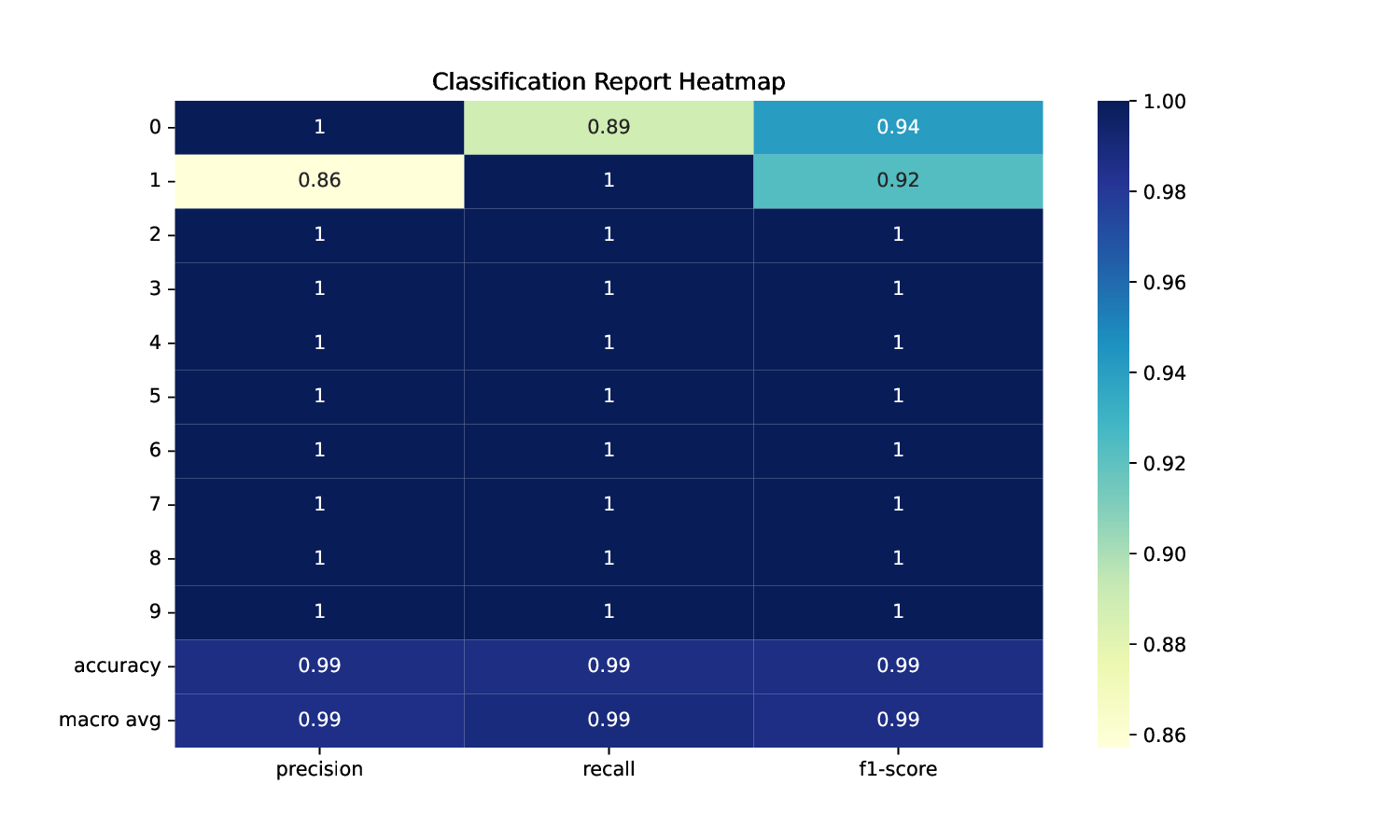}
        \caption{Heatmaps on Dataset C}
        \label{fig:datasetC}
    \end{subfigure}
    
    \caption{Heatmaps of the classification report for the proposed model on three datasets: (a) Dataset A, (b) Dataset B, and (c) Dataset C. Each heatmap displays the Precision, Recall, and F1-Score for the 10 different classes. Source: Author's own.}
    \label{fig:heatmapIndi}
\end{figure}

\noindent The consistently high scores across all datasets underline the model’s reliability and robustness in real-world predictive maintenance tasks, where both precision and recall are crucial for efficiently detecting bearing system faults. Figure \ref{fig:heatmapIndi} presents heatmaps of the classification reports for each dataset, illustrating the Precision, Recall, and F1-Score per class. On Dataset A, the model performs exceptionally well, with minor reductions in Recall for class 3 (0.92) and in F1-Score for class 7 (0.91), while still maintaining an overall accuracy of 99\%. Dataset B achieves perfect classification across all metrics and classes. For Dataset C, the model again performs strongly, with slight decreases in Recall for class 1 (0.89) and class 3 (0.92), maintaining an overall accuracy of 99\%.

\noindent The proposed ATBTSGM model is compared against several baseline models, including: (1) GNNBFD \cite{xiao2023graph}, (2) Neural Network-based SO-GAAL (Single Objective Generative Adversarial Active Learning), (3) CutPC (a graph-based clustering method using noise cutting), (4) LOF (Local Outlier Factor), (5) distance-based methods including KNN, and (6) IForest, as mentioned in \cite{xiao2023graph}. These comparisons are conducted across three datasets (A, B, and C). Key evaluation metrics such as Accuracy (ACC), Detection Rate (DR), False Alarm Rate (FAR), and Area Under the Curve (AUC) are used to compare performance, as summarized in Tables \ref{tab:acc_dr_values} and \ref{tab:far_auc_values}.

\begin{table}[ht!]
    \centering
    \caption{ACC and DR values for different models across datasets. Source: Author's own.}
    \begin{tabular}{@{}lcccccc@{}}
        \toprule
        \multirow{2}{*}{Model} & \multicolumn{3}{c}{ACC (\%)} & \multicolumn{3}{c}{DR (\%)} \\ 
        \cmidrule(lr){2-4} \cmidrule(lr){5-7}
                               & Dataset A & Dataset B & Dataset C & Dataset A & Dataset B & Dataset C \\ \midrule
        GNNBFD                 & 99.30     & 98.60     & 99.30     & 95.0      & 90.0      & 95.0      \\ \midrule
        SO-GAAL                & 96.51     & 96.51     & 97.21     & 75.0      & 75.0      & 80.0      \\ \midrule
        CutPC                  & 97.21     & 95.81     & 95.81     & 80.0      & 70.0      & 70.0      \\ \midrule
        LOF                    & 93.72     & 92.79     & 96.28     & 55.0      & 40.0      & 73.33     \\ \midrule
        KNN                    & 93.02     & 91.05     & 95.12     & 50.0      & 35.0      & 65.0      \\ \midrule
        IForest                & 97.21     & 97.21     & 97.91     & 80.0      & 80.0      & 85.0      \\ \midrule
        ATBTSGM                & 99.40     & 100.0     & 99.40     & 98.0      & 97.5      & 96.3      \\ \bottomrule
    \end{tabular}
    \label{tab:acc_dr_values}
\end{table}

\begin{table}[ht!]
    \centering
    \caption{FAR and AUC values for different models across datasets. Source: Author's own.}
    \begin{tabular}{@{}lcccccc@{}}
        \toprule
        \multirow{2}{*}{Model} & \multicolumn{3}{c}{FAR} & \multicolumn{3}{c}{AUC} \\ 
        \cmidrule(lr){2-4} \cmidrule(lr){5-7}
                               & Dataset A & Dataset B & Dataset C & Dataset A & Dataset B & Dataset C \\ \midrule
        GNNBFD                 & 0.375     & 0.750     & 0.375     & 0.983     & 0.967     & 0.991     \\ \midrule
        SO-GAAL                & 1.875     & 1.875     & 1.500     & 0.894     & 0.863     & 0.955     \\ \midrule
        CutPC                  & 1.500     & 4.500     & 2.250     & 0.962     & 0.814     & 0.909     \\ \midrule
        LOF                    & 3.375     & 2.250     & 2.000     & 0.710     & 0.735     & 0.946     \\ \midrule
        KNN                    & 3.750     & 4.875     & 2.625     & 0.658     & 0.673     & 0.8902    \\ \midrule
        IForest                & 1.500     & 1.500     & 1.125     & 0.923     & 0.874     & 0.966     \\ \midrule
        ATBTSGM                & 0.100     & 0.100     & 0.200     & 1.000     & 1.000     & 0.990     \\ \bottomrule
    \end{tabular}
    \label{tab:far_auc_values}
\end{table}

\noindent The ATBTSGM model demonstrated exceptional performance across all datasets, achieving the highest ACC (99.40\% for Datasets A and C, and 100\% for Dataset B). Its detection rates (DR) were equally impressive, ranging from 96.3\% to 98\%. Furthermore, the model maintained a very low False Alarm Rate (FAR), with values as low as 0.1 for Datasets A and B, and 0.2 for Dataset C. ATBTSGM also attained near-perfect AUC values, reinforcing its reliability and superior performance when compared to other baseline models.

\noindent In contrast, GNNBFD, though performing well, showed slightly lower ACC and DR, and a higher FAR (up to 0.750 for Dataset B). Other models, such as SO-GAAL and CutPC, performed moderately, while LOF and KNN struggled, particularly on Dataset B, with low DR and high FAR.

\noindent Overall, the ATBTSGM model consistently outperformed its competitors across all metrics, demonstrating its robustness, low false alarm rates, and high fault detection capabilities, making it an optimal choice for predictive maintenance applications.

\begin{figure}[h!]

    \centering

    \includegraphics[width=0.7\linewidth]{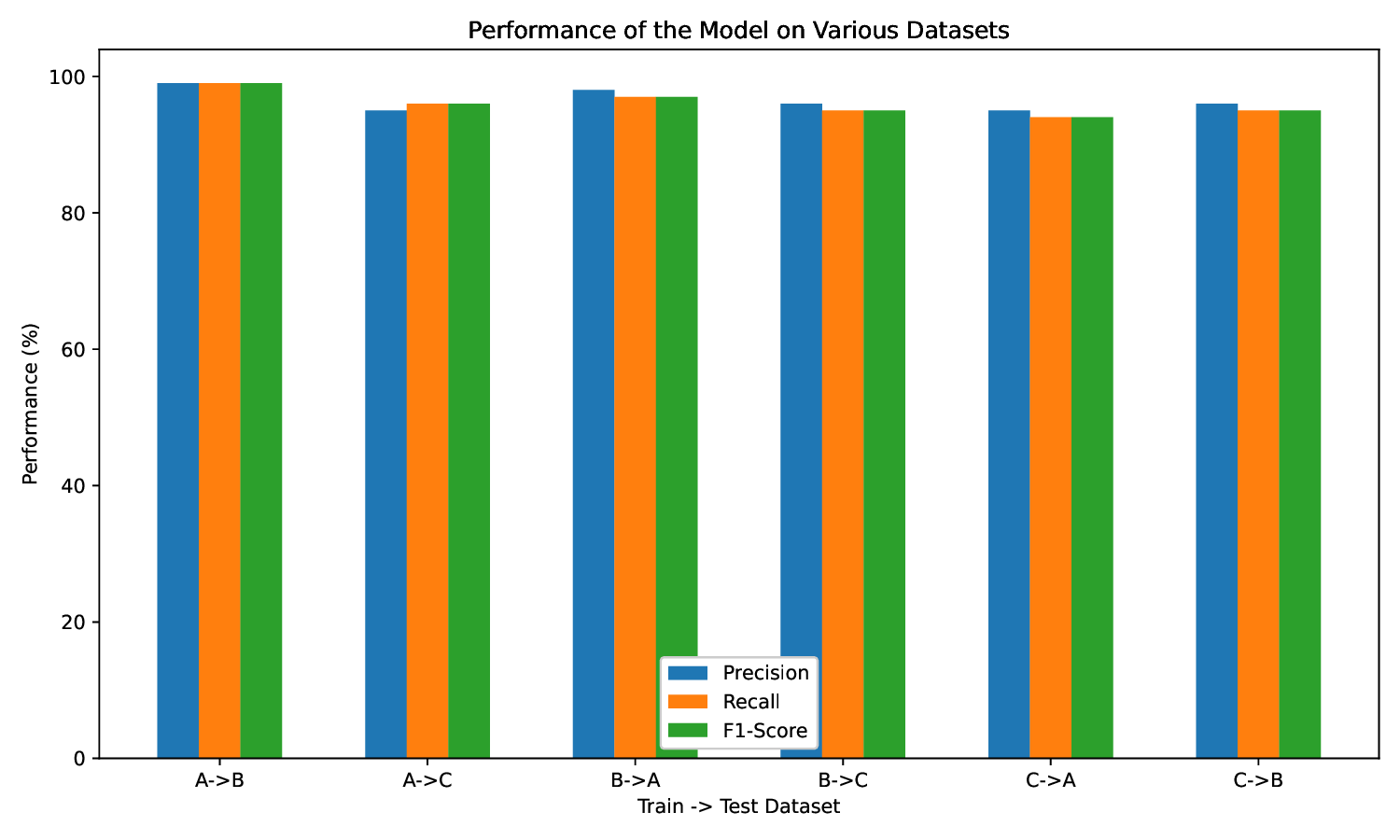}

    \caption{Performance of ATBTSGM on Various Datasets. Source: Author's own.}
    \label{fig:PerformanceofModel}
\end{figure}

\noindent The generalization ability of the proposed ATBTSGM model was evaluated by training and testing it on different combinations of Datasets A, B, and C, using Precision, Recall, and F1-Score as key performance metrics. Figure \ref{fig:PerformanceofModel} presents the model’s performance, demonstrating its robustness across different dataset configurations. For instance, training on Dataset A and testing on Dataset B yields near-perfect results, with 99\% achieved across all three metrics. However, when tested on Dataset C, a slight performance decrease is observed, with Precision at 95\%, Recall at 96\%, and F1-Score at 96\%.

\noindent Similarly, when trained on Dataset B, the model maintains strong performance when tested on Dataset A, achieving 98\% Precision, 97\% Recall, and 97\% F1-Score. Testing on Dataset C results in a minor decline, with all metrics around 95\%. Training the model on Dataset C delivers consistent results when tested on both Datasets A and B, with metrics consistently near 95\%. This cross-dataset evaluation highlights the model’s ability to generalize across varying data sources, a critical feature for real-world applications, where data often vary in structure and noise levels.


\begin{figure}[!htbp]

    \centering

    \includegraphics[width=0.7\linewidth]{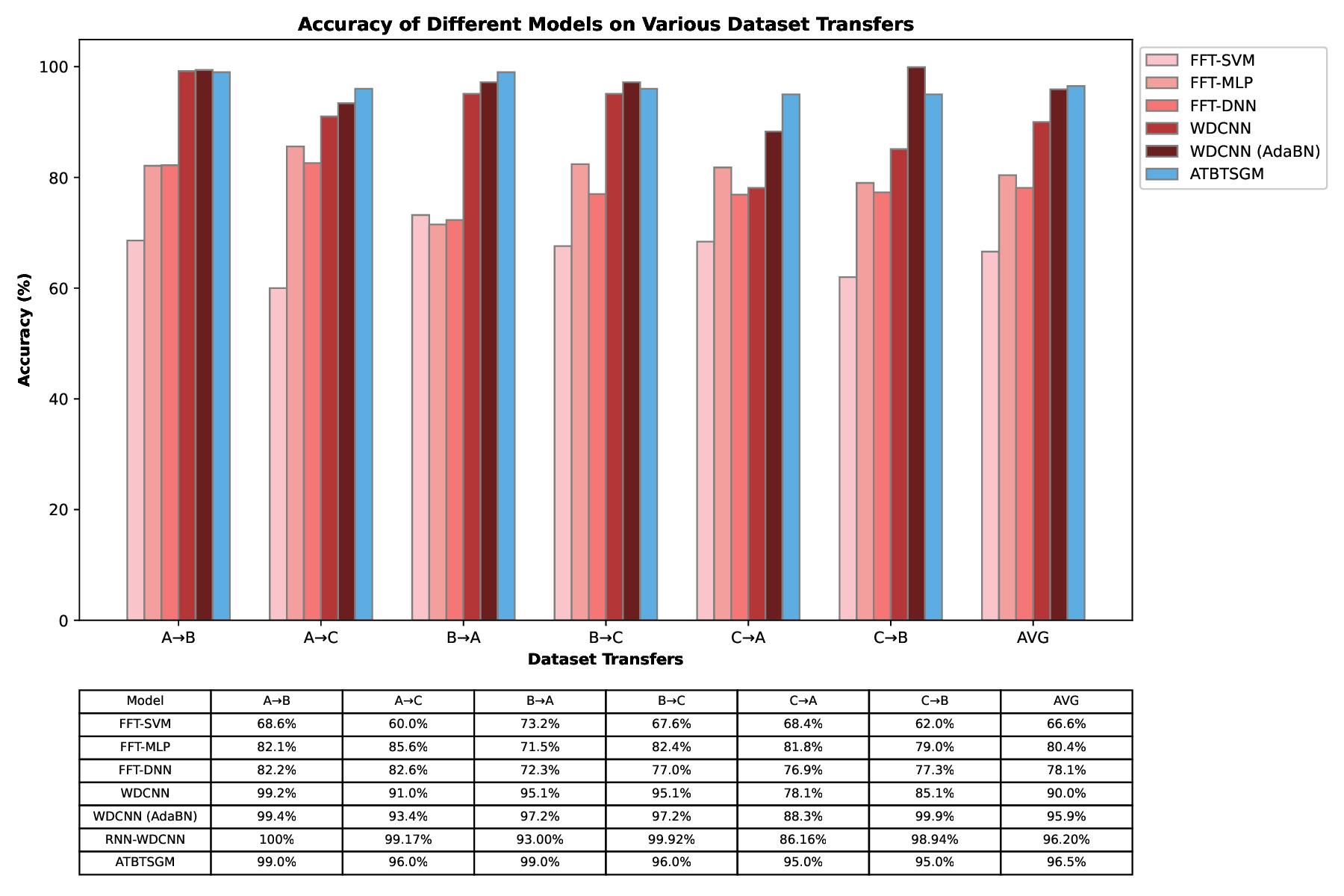}

    \caption{Accuracy of Different Models on Various Dataset Transfers, Including Proposed Model. Source: Author's own.}
    \label{fig:performance}
\end{figure}

\noindent Cross-dataset testing underscores the model’s strong generalization abilities, particularly when trained on Datasets A and B. The performance remains close to perfect on similar datasets but drops slightly when tested on Dataset C, indicating that Dataset C may contain greater variability or complexity. This suggests potential areas for model enhancement, particularly in handling more diverse datasets.

\noindent Figure \ref{fig:performance} compares the ATBTSGM model’s accuracy in various dataset transfer scenarios against baseline models. The FFT-SVM model \cite{amar2014vibration}, for example, achieves an average accuracy of 66.6\%, while FFT-MLP \cite{saravanan2010incipient} and FFT-DNN \cite{janssens2016convolutional} models show improved accuracy at 80.4\% and 78.1\%, respectively. The WDCNN \cite{zhang2017new} model performs well, with an average accuracy of 90.0\%, which is further improved to 95.9\% with the WDCNN (AdaBN) model \cite{zhang2017new}. The RNN-WDCNN \cite{shenfield2020novel} model boosts performance further, reaching 96.2\% on average. Finally, the ATBTSGM model surpasses all other models, achieving an average accuracy of 96.5\%, with individual results ranging from 95.0\% to 99.0\% across dataset transfers.

\section{Discussion}

The superior performance of ATBTSGM can be attributed to two key factors: its graph-based data representation that incorporates entropy and its capability to capture temporal dependencies within time series data.

\noindent ATBTSGM employs a graph-based representation of the data, leveraging entropy to evaluate uncertainty and complexity in the dataset. Entropy acts as a measure of the informational content within the data, enabling the model to focus on the most relevant and meaningful features. This approach allows the model to uncover subtle and intricate patterns, particularly effective for fault detection tasks where hidden relationships between features are critical. By capturing these relationships, the model can perform a more sophisticated analysis of the data than traditional machine learning methods that treat data as independent points.

\noindent In addition, the model excels at capturing temporal dependencies within time series data. Time series data often involve sequential correlations, where each data point is influenced by previous values. By leveraging the graph structure, ATBTSGM effectively incorporates this temporal context, crucial for understanding the dynamic behavior in systems like fault detection. This ensures that the model retains essential contextual information, leading to more accurate predictions.

\noindent These two factors work together to significantly enhance ATBTSGM’s performance. Its ability to capture both temporal relationships and graph-based structural patterns results in high accuracy, improved detection rates, and a reduced false alarm rate across all datasets. This combination makes ATBTSGM a robust and reliable solution for real-world predictive maintenance and fault detection applications.

\section{Conclusion}

Across multiple datasets, the ATBTSGM model consistently demonstrates exceptional performance in fault detection, with high Precision, Recall, and F1-Score metrics. With exceptional accuracy on Datasets A and C (99\%) and impeccable performance on Dataset B (100\%), the model has shown its reliability and resilience in identifying faults in diverse scenarios. The heatmaps further validate the model’s strong classification ability in 10 different classes, with only slight performance decreases in certain cases. Additionally, the model’s superior performance in real-world predictive maintenance tasks is highlighted by its low False Alarm Rate (FAR) and high Detection Rate (DR) across all datasets. In addition to surpassing baseline models in accuracy and generalization, the ATBTSGM model also maintains strong fault detection capabilities, low error rates, and high resilience to varying data characteristics. By offering both high accuracy and reliable predictive maintenance, the ATBTSGM model is established as the optimal choice for fault detection in complex mechanical systems.

\noindent Despite the impressive results of the ATBTSGM model on diverse datasets, there is potential for future research to improve its scalability for larger and more intricate datasets, considering the substantial amount of data generated by real-world systems. Moreover, the inclusion of domain adaptation and transfer learning techniques enables the model to adjust to diverse operating conditions without extensive retraining. Implementing Transformer-based architectures could further amplify the model’s capability to capture long-term dependencies in its temporal representation. A potential area for future research is the implementation of real-time fault detection with reduced computational latency. Furthermore, it is important to note that the issue of robustness to noisy or incomplete data, commonly faced in industrial applications, requires further consideration. Enhancing the model’s explainability and interpretability is vital for establishing trust in high-stakes predictive maintenance scenarios, while mitigating computational complexity and generalizing to unseen fault types are necessary for practical deployment.

\section*{Statements and Declarations}

\subsection*{Competing Interests}
The authors declare that there are no competing interests associated with this research work.

\subsection*{Funding}
This research did not receive any specific grant from funding agencies in the public, commercial, or not-for-profit sectors.

\subsection*{Informed Consent}
Informed consent was obtained from all individual participants included in the study.

\subsection*{Data Availability}
The datasets generated and/or analyzed during the current study are available in upon reasonable request from the corresponding author.

\bibliographystyle{ieeetr}
\bibliography{main}

\end{document}